\documentclass{article}
\pdfoutput=1

\usepackage[preprint]{neurips_2022}

\usepackage[utf8]{inputenc} 
\usepackage[T1]{fontenc}    
\usepackage{hyperref}       
\usepackage{url}            
\usepackage{booktabs}       
\usepackage{bookmark}
\usepackage{amsfonts}       
\usepackage{nicefrac}       
\usepackage{microtype}      
\usepackage{xcolor}         
\usepackage{amsmath}
\usepackage{graphicx}
\usepackage{bbm}

\title{Optimizing Industrial HVAC Systems with Hierarchical Reinforcement Learning}

\author{
  William Wong\thanks{Work done while at DeepMind.}\\
  Carnegie Mellon University\\
  \texttt{w07wong@gmail.com} \\
   \And
   Praneet Dutta \\
   DeepMind \\
   \texttt{praneetdutta@deepmind.com} \\
   \AND
   Octavian Voicu \\
   DeepMind \\
   \texttt{octav@deepmind.com} \\
   \And
   Yuri Chervonyi \\
   DeepMind \\
   \texttt{cyuri@deepmind.com} \\
   \And
   Cosmin Paduraru \\
   DeepMind \\
   \texttt{paduraru@deepmind.com} \\
   \And
   Jerry Luo \\
   DeepMind \\
   \texttt{jerryjluo@deepmind.com} \\
}

\begin{document}

\maketitle

\begin{abstract}
Reinforcement learning (RL) techniques have been developed to optimize industrial cooling systems, offering substantial energy savings compared to traditional heuristic policies. A major challenge in industrial control involves learning behaviors that are feasible in the real world due to machinery constraints. For example, certain actions can only be executed every few hours while other actions can be taken more frequently. Without extensive reward engineering and experimentation, an RL agent may not learn realistic operation of machinery.To address this, we use hierarchical reinforcement learning with multiple agents that control subsets of actions according to their operation time scales.Our hierarchical approach achieves energy savings over existing baselines while maintaining constraints such as operating chillers within safe bounds in a simulated HVAC control environment.
\end{abstract}

\section{Introduction}
Industrial systems account for 54\% of global energy usage and 34\% of greenhouse gas emissions \citep{industrial-sector-energy-consumption, portner2022climate}. With energy consumption increasing by 1\% yearly, industrial systems play an enormous role in rising global temperature \citep{energy-increase-report}. One such industrial system is a HVAC (heating, ventilation, and air conditioning) system. This consists of machinery which regulates temperature in data centers, offices, commercial buildings etc. Specifically, HVAC cooling systems by themselves account for 10\% of greenhouse gas emissions \citep{tessler2017deep}.

Traditionally, controllers must be tuned for an environment and their performances degrade when operating conditions change \citep{afram2014theory}. Furthermore, hand tuning a controller to minimize energy usage and keep the temperature within certain constraints can be challenging.

Instead, reinforcement learning can aid operators by acting as a supervisory controller which determines setpoints for controllers to meet. By posing energy savings and temperature constraints as an optimization problem, RL can determine more efficient setpoints. For instance, \cite{google-ai-data-center} reduced data center cooling energy usage by 40\% with RL. However, applying a learned policy to a real-life system poses many challenges. For one, an agent may learn to turn HVAC equipment on and off frequently, or leave them on for extended periods of time. In the real-world, building operators avoid this behavior to limit wear and tear. For offline RL, regularized behavior value estimation \citep{gulcehre2021regularized} can prevent an agent from generating unrealistic behavior not seen in production, but the core issue is the difficulty for a single agent to reason across both extremely long and short time horizons. Instead, we propose using multiple agents, each operating at different timescales, to address this issue.

We focus on optimizing chiller plants, a component of HVAC systems. These plants consist of multiple chillers, mechanical devices that are responsible for removing heat from the buildings typically via a liquid refrigerant. Chillers should only be turned on and off every few hours and usage should be spread equally among chillers to avoid unnecessary wear and tear. At the same time, building temperature needs to be maintained within specified bounds throughout chiller cycling. Hierarchical reinforcement learning (HRL) offers the ability to reason across different time scales. We propose an HRL approach which avoids the necessity of extensive reward engineering to meet building temperature requriements and minimize chiller wear and tear. While our work focuses on HVAC systems, the same methods can be applied to other industrial systems.

We validate our approach in a high-fidelity industrial cooling system simulator \citep{chervonyi2022semi} whose simulations have been verified against real-world industrial system data, parameterized for re-creating various scenarios. Additionally, simulation steps take on average 20 to 40 seconds. As a result, simulating and training on a large number of rollouts is difficult. Similar to using real-world offline data, agents must be sample efficient and learn with a limited amount of data.

In summary, our contributions are: \textbf{(1)} we present a novel hierarchical reinforcement learning architecture for optimizing chiller plants. \textbf{(2)} For comparison, we develop a baseline heuristic based policy (HBP) and a mulit-agent RL (MARL) approach. Our heuristic policy is inspired by real heuristic policies found in real buildings. \textbf{(3)} We demonstrate that our hierarchical architecture outperforms flat RL algorithms, HBP, and MARL on a task with competing objectives over long time horizons. This suggests hierarchical reinforcement learning as a framework to achieve desired real-world performance for HVAC control.

\section{Related Work}
There has been considerable work optimizing HVAC systems using machine learning. In the online setting, \citet{en12152860, panahizadeh2021evaluation, park2019machine} have leveraged neural networks to build data driven models of chillers which can be used downstream for optimization problems. 

Discovering optimized strategies have largely centered around reinforcement learning based approaches. Policy gradient methods such as PPO \citep{schulman2017proximal}, DDPG \citep{lillicrap2015continuous}, and A3C \citep{mnih2016asynchronous} have generated more efficient control mechanisms \citep{gao2019energy, azuatalam2020reinforcement, zhang2019whole, du2021intelligent}. For example, \citet{du2021intelligent} demonstrated how DDPG can reduce energy consumption by 15\% and reduce comfort violations (how much temperature strays from specified bounds) by 79\% compared to DQN for multi-zone residential HVAC control. Q-learning methods have also been applied, offering 20-70\% energy reductions from rule-based baseline strategies \citep{wei2017deep, raman2020reinforcement}. Our work differs from these by introducing multiple agents to control different components of an HVAC system.

\citet{chervonyi2022semi} explores DMPO \citep{hoffman2020acme}, DDPG, and D4PG \citep{barth2018distributed} for industrial cooling  control. We expand on this by exploring hierarchical and multi-agent reinforcement learning. \citet{hanumaiah2021distributed} explore a multi-agent approach to HVAC control. The authors propose adjusting temperature setpoints in order to minimize energy usage while maximizing temperature comfort. Our approach optimizes for energy usage and temperature comfort as well, while additionally introducing an entropy based objective to balance chiller utilization.

To the best of our knowledge, hierarchical reinforcement learning has not been explored for HVAC control. Hierarchical frameworks were introduced with the goal of enabling agents to perform long, complex tasks which can be decomposed into simpler sub-tasks that operate at different timescales \citep{sutton1999between, precup2000temporal}. A popular hierarchical formulation is the options framework where a top level iteratively selects options, or sub-policies, which act until a termination criteria is met. \citet{vezhnevets2017feudal} extends this option framework by having a high level agent (HLA) provide an explicit goal for a low level agent (LLA) to follow. Our approach is similar with our HLA invoking the LLA for some number of steps which is visible to the LLA as a goal. As described in Section~\ref{sec:hrl}, our high level agent can also directly interact with the environment. In line with this, \citet{tessler2017deep} utilize HRL for Minecraft, allowing a high level agent to either invoke pre-learned sequences of primitive actions represented as deep networks, or primitive actions directly.

\section{Preliminaries}

\subsection{Industrial Cooling Systems}
\begin{figure}[t!]
    \centering
    \includegraphics[width=6.67cm, height=5cm]{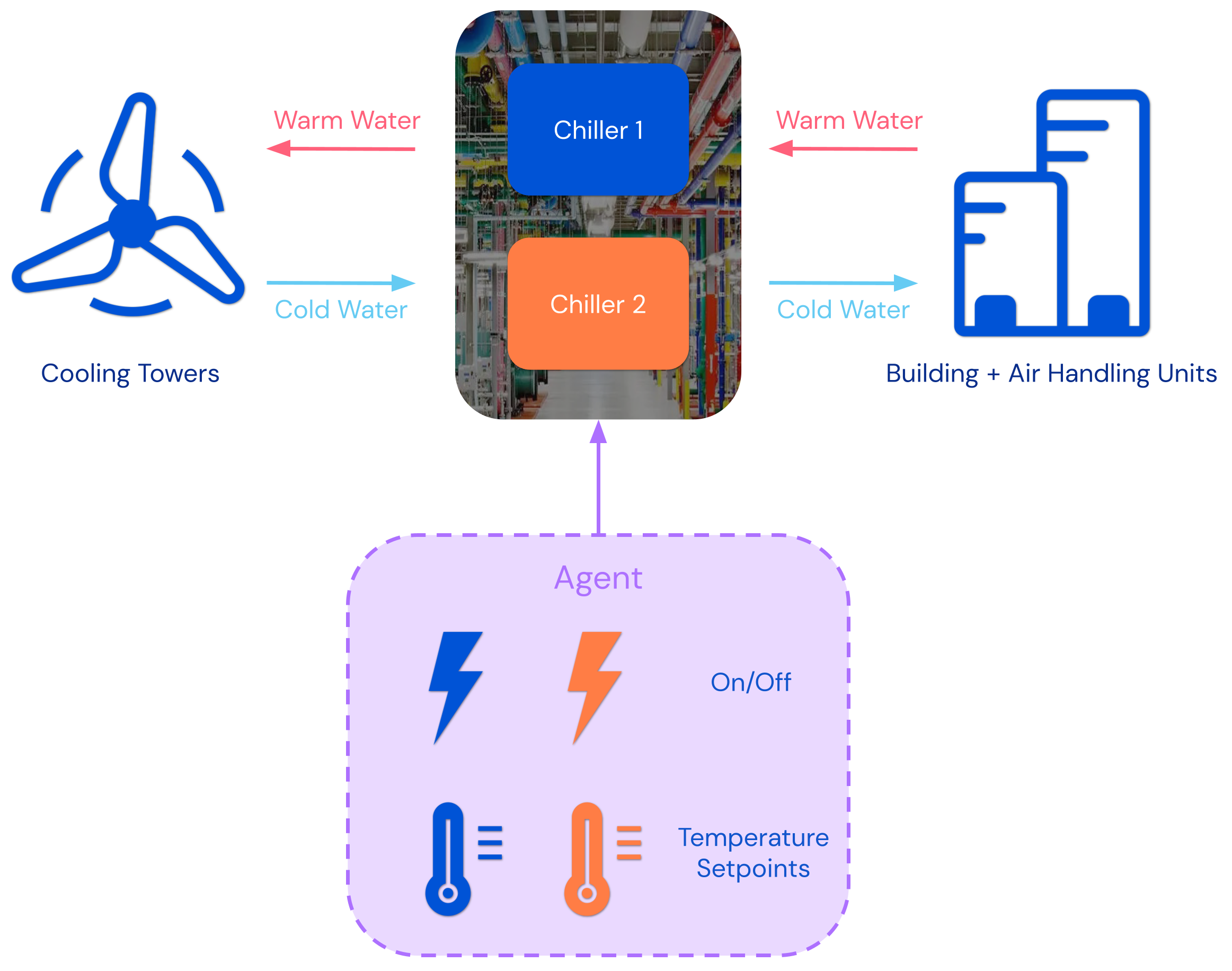}
    \caption{Agent interaction with an HVAC system. We allow agents to enable chillers and adjust their temperature setpoints. Chillers regulate the cold water temperature used to cool building air.}
    \label{fig:environment_loop}
\end{figure}

We refer the reader to \cite{chervonyi2022semi} for a complete overview of industrial cooling systems. Throughout the day, solar radiation and building occupants (people, computers, etc.) generate heat and warm a building's air. In order to keep building temperature at a comfortable level, the warm air is cooled by cold water provided by chillers. This heat exchange between cold water and warm air causes the water to heat up. The role of chillers is to cool the warmed water down to a certain temperature specified by a control called a temperature setpoint. Once cooled, water returns to a building where PID controllers use the water to meet temperature setpoints inside the building. The colder the water, the easier it is to achieve those setpoints and vise versa. We allow agents to turn chillers on and off, and adjust chiller temperature setpoints to control building temperature (Figure~\ref{fig:environment_loop}).

\subsection{Problem Setup}
\label{sec:problem_setup}
We formulate controlling chiller plants as an MDP \citep{sutton2018reinforcement}. Steps in the environment correspond to five minutes. At each step $t$, an agent provides an action vector $\textbf{a} \in \mathbb{R}^{2n_\text{tot}}$ where $n_\text{tot}$ is the total number of chillers. Each chiller can be switched on and have their temperature setpoint adjusted. The state $\textbf{s}(t)$ includes chiller plant measurements including, but not limited to, the number of chillers enabled, building temperature, and chiller power. To minimize energy usage, meet building temperature constraints, and balance the usage of each chiller, the reward function is
$\textbf{R}(t) = \alpha_{h} \cdot \textbf{h}(t)^{\lambda_{h}} - \alpha_{o} \cdot \mathbbm{1}(n_e \neq n_d) + \alpha_{p} \cdot \textbf{p}(t)^{\lambda_{p}} - \alpha_{c} \cdot \textbf{c}(t)^{\lambda_{c}}$ with all $\alpha > 0$. To balance load across all chillers, we compute the normalized entropy $\textbf{h}(t) = \frac{1}{\log n_\text{tot}} \sum_{i}^{n_\text{tot}} -p_i \log p_i$ where $p_i = \frac{\text{chiller i on time}}{\sum_{j \neq i}^{n_\text{tot}} \text{chiller j on time}}$. Balancing chiller usage is trivial if the agent turns all chillers on or off all the time, so we add a penalty $-\alpha_{o} \mathbbm{1}(n_e \neq n_d)$ for when the number of chillers enabled $n_e$ is not the desired amount of chillers $n_d$ that should be on at all times. The value $n_d$ chillers need to be able to meet temperature requirements and is predetermined by building operators. To minimize energy, we reward agents $\textbf{p}(t) = (\frac{w(t)}{1000} + 1)^{-1}$ where $w(t)$ is the amount of power used at step $t$. Finally, we minimize temperature violations with $\textbf{c}(t) = \max(v_\text{upper}(t), v_\text{lower}(t))$ where $v_\text{upper}(t)$ and $v_\text{lower}(t)$ are how much the dry bulb temperature violates temperature upper and lower bounds respectively.
\section{Hierarchical Reinforcement Learning}
\label{sec:hrl}

\subsection{Motivation}
As shown in Section \ref{sec:experiments}, RL-based agents turn chillers on and off frequently to balance chiller usage while meeting temperature constraints. However, this is costly energy-wise and mechanically as chillers are physical machines which undergo wear and tear. To prevent this, enabling chillers should be delegated to an agent which acts infrequently while another agent, which acts more frequently, adjusts temperature setpoints to maintain building temperature. Determining how frequent each agent should act can be fixed or learned. A fixed frequency corresponds to a multi-agent reinforcement learning setup with two agents taking turns controlling setpoints. However, the optimal control frequency is difficult to determine and is limiting in dynamic environments where building requirements call for different action lengths. Instead, the control frequency can be learned with hierarchical reinforcement learning where a high level agent determines when to update chiller enabled states.

\subsection{Option Framework}
We explore a two level hierarchy which is commonly formulated with the options framework \citep{sutton1999between}. An option is a temporally extended action with three components: a policy $\pi_o$, a termination condition $\beta$, and an initiation set $\mathcal{I}$. At timestep $t$, a high level agent picks an option among those where the current state $\textbf{s}(t) \in \mathcal{I}$. The option executes until a termination condition is met. Afterwards, the high level agent selects a new option. In our work, we delegate learning and selecting options to a low level agent. The high level agent provides a termination condition which can be thought of a goal the option needs to achieve.

\subsection{Architecture}
\begin{figure}[t!]
    \centering
    \includegraphics[width=7cm, height=5cm]{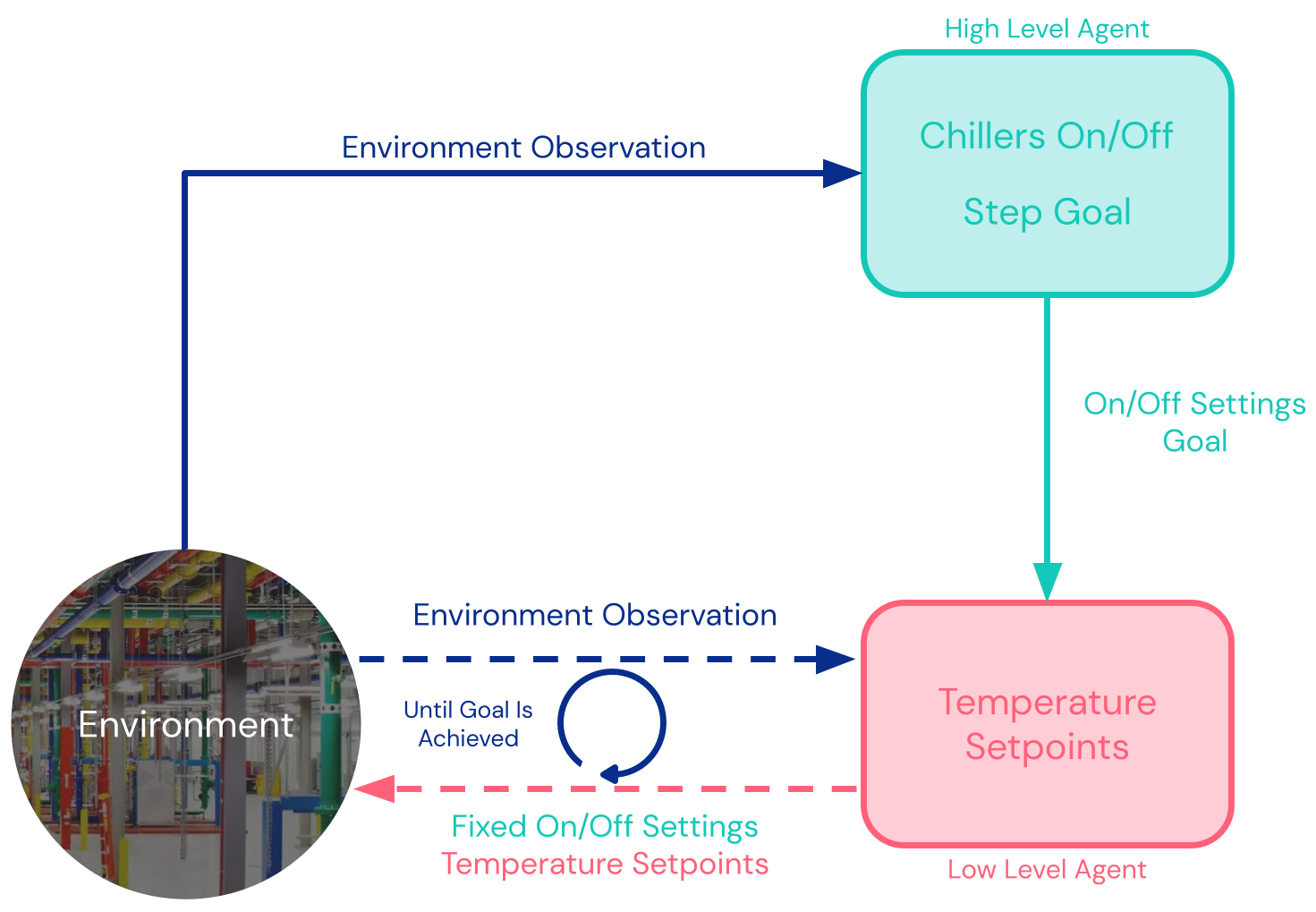} 
    \caption{Hierarchical reinforcement learning agent loop. The high level agent determines which chillers to turn on and invokes the low level agent for some number of steps.}
    \label{fig:hrl_architecture}
\end{figure}

Our hierarchical reinforcement learning \citep{vezhnevets2017feudal} model consists of two agents: a high level agent (HLA) which turns chillers on and off, and a low level agent (LLA) which manipulates temperature setpoints (Figure \ref{fig:hrl_architecture}). At every timestep, the high level agent can choose to change the enabled state of chillers, or provide a step goal - the number of steps to invoke the low level agent for - ranging from 5 minutes to 4 hours. To have the high level agent learn to balance chiller usage while minimizing energy consumption, we provide the reward $\textbf{R}_{\text{HLA}}(t) = \alpha_{h} \cdot \textbf{h}(t)^{\lambda_{h}} - \alpha_{o} \cdot \mathbbm{1}(n_e \neq n_d) + \alpha_{p} \cdot \textbf{p}(t)^{\lambda_{p}}$. The power term prevents the agent from cycling chillers frequently by penalizing energy wastage during chiller start-up. The low level agent is tasked with meeting temperature constraints and is rewarded $\textbf{R}_{\text{LLA}}(t) = \alpha_{p} \cdot \textbf{p}(t)^{\lambda_{p}} + \alpha_{c} \cdot \textbf{c}(t)^{\lambda_{c}}$. The high level agent's observation space is the environment observation. The low level agent observes the environment observation and the high level agent's action and step goal.

\section{Experiments}
\label{sec:experiments}
To demonstrate how HRL can learn safer and more efficient chiller sequences compared to flat RL methods, we define a task configuration leveraging the simulator introduced in \cite{chervonyi2022semi}. We simulate a real world industrial facility where the building load fluctuates throughout the day. This dictates the cooling requirements from the chiller plant and is a weighted function of the dry bulb temperature and building heat. Building heat can be approximated by the number of occupants in the building, for example, and is represented by the flow of hot fluid, "load inlet velocity", through air handling units.

\subsection{Baseline Methods}
We compare our hierarchical agent against RL algorithms DMPO, DDPG, and D4PG, an HBP policy, and a multi-agent RL approach. We create an HBP policy which uses building temperature to determine how many chillers to run. If the building temperature is greater than the upper temperature constraint for 10 minutes, an additional chiller is turned on. If the building temperature is lower than the bottom temperature constraint, and this holds true for 15 minutes, a chiller is turned off. While an optimal HBP policy can be created for a specific environment, finding such a policy is infeasible without a simulator for frequent experimentation or significant knowledge of a facility. This complexity scales as we increase the dimensionality of the action and control space. Additionally, since HBP is a fixed policy, it cannot adapt to new situations without re-tuning.

The MARL policy features two agents: a high level agent which turns chillers on and off every 1 hour and a low level agent which controls temperature setpoints when the high level agent is not acting. Similar to the HRL agent, the MARL agent splits the reward function between its two agents. The high level agent is rewarded $\textbf{R}_{\text{HLA}}(t) = \alpha_{h} \cdot \textbf{h}(t)^{\lambda_{h}} - \alpha_{o} \cdot \mathbbm{1}(n_e \neq n_d) + \alpha_{p} \cdot \textbf{p}(t)^{\lambda_{p}}$. The low level agent receives $\textbf{R}_{\text{LLA}}(t) = \alpha_{p} \cdot \textbf{p}(t)^{\lambda_{p}} + \alpha_{c} \cdot \textbf{c}(t)^{\lambda_{c}}$.

\subsection{Experiment Setup}
\label{sec:experiment_setup}
Experiment use episodes simulating half a day, 12 hours. To mimic a scenario where building load oscillates during a day (for example, office workers go to work then leave), we model load inlet velocity with a sine wave having a fixed amplitude of 2 m/s and a period of 400 minutes. Dry bulb temperature can have increased variance  due to the  weather. We model this with a sine wave having a random amplitude from 1 to 10 degrees Fahrenheit and a period of 200 minutes. For evaluation, we set the lower and upper temperature constraints to be 50 and 60 degrees Fahrenheit respectively. RL agents can control two chillers with temperature setpoints in the range of 38 to 46 degrees Fahrenheit. The HBP has a fixed temperature setpoint of 41 degrees Fahrenheit. We setup the environment so that one chiller needs to be on at all times (i.e. $n_d = 1$). If no chillers are on, the building temperature will violate the upper temperature constraint.

Both the high level and low level agents for MARL and HRL use DMPO for optimization. We use the ACME framework \citep{hoffman2020acme} to create and train all agents. DMPO-based agents use an epsilon of 1e-1 and epsilon mean of 1e-2. We use the standard hyperparameters from ACME for DDPG and D4PG. We used approximately 14 million vCPU core hours on Google Cloud Platform.

\subsection{Reward Discussion}
\label{sec:reward-discussion}
Many real world problems are multi-objective, making defining a single reward function to be difficult. We desired a policy which uses one chiller at all times (to minimize energy usage), switches between chillers every few hours (for equal utilization), and meets building temperature requirements. We experimented with many reward functions including penalizing frequent chiller cycling, but these rewards were difficult for agents to learn any meaningful behavior. In general, it was difficult to specify a reward function that captured user preferences. However as we will see in Section~\ref{sec:results}, HRL removed the need for complex reward engineering to achieve the desiderata. Through a grid search, we found that $\alpha_{h} = 30, \lambda_{h} = 5, \alpha_{o} = 25, \alpha_{p} = 4, \lambda_{p} = 2, \alpha_{c} = 2,$ and $\lambda_{c} = 2$ promoted chiller switching with DMPO. We applied the same parameters to all other methods. To meet temperature constraints, we found that $v_\text{upper}(t) = 57$ and $v_\text{lower}(t) = 53$ work best. This is due to the large variance in building load and the capabilities of the chillers.

\subsection{Results}
\label{sec:results}

\begin{figure}[t!]
    \centering
    \includegraphics[width=90mm]{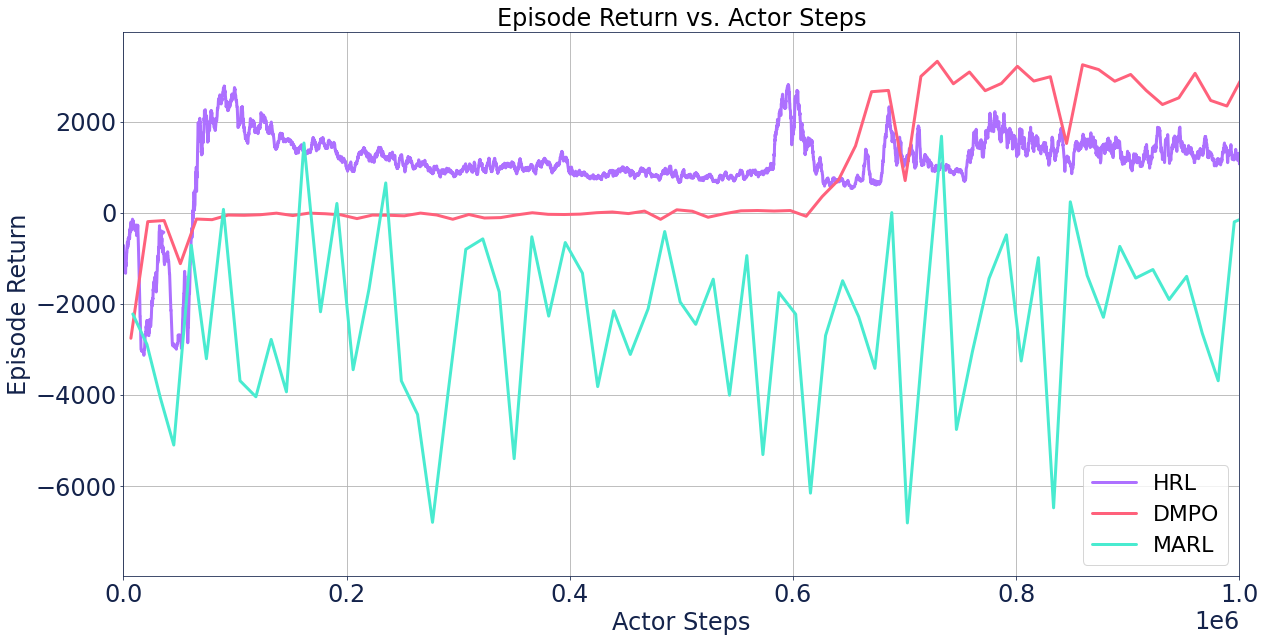}
    \caption{Episode return during first 1 million steps of training for DMPO, multi-agent RL and hierarchical RL agents. HRL is more sample efficient than DMPO, achieving higher reward within the first 100,000 steps. While DMPO converges to a higher episode return, this is due to DMPO manipulating chillers in an unsafe way to maximize the task balance reward. We constructed the reward function to see if \textit{any} agent could learn safe and efficient chiller sequences and are more concerned with downstream behavior than episode return.}
    \label{fig:episode_reward}
\end{figure}

\begin{figure}[htb!]
    \centering
    \begin{tabular}{cc}
        \includegraphics[width=65mm]{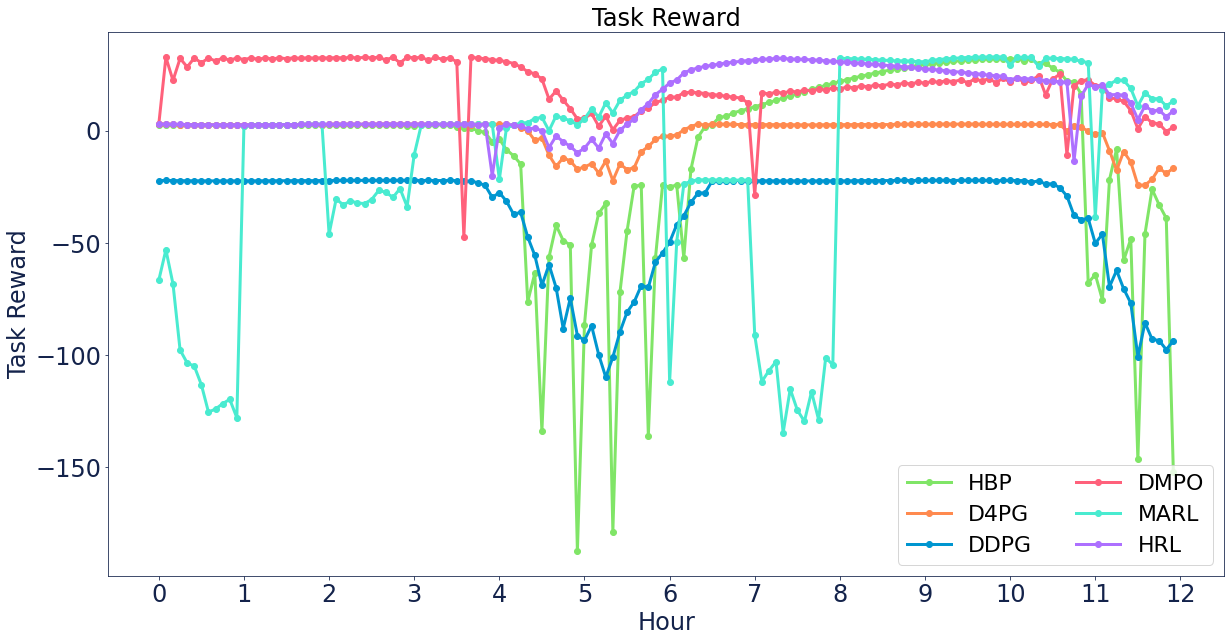} &  \includegraphics[width=65mm]{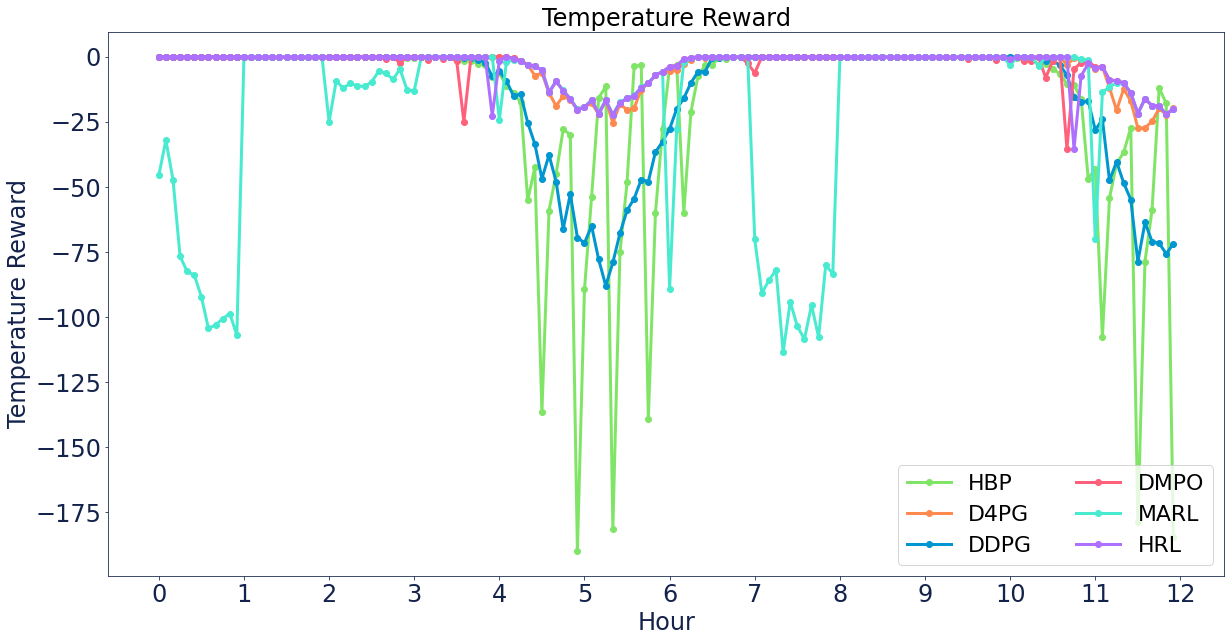}\\
        
        \includegraphics[width=65mm]{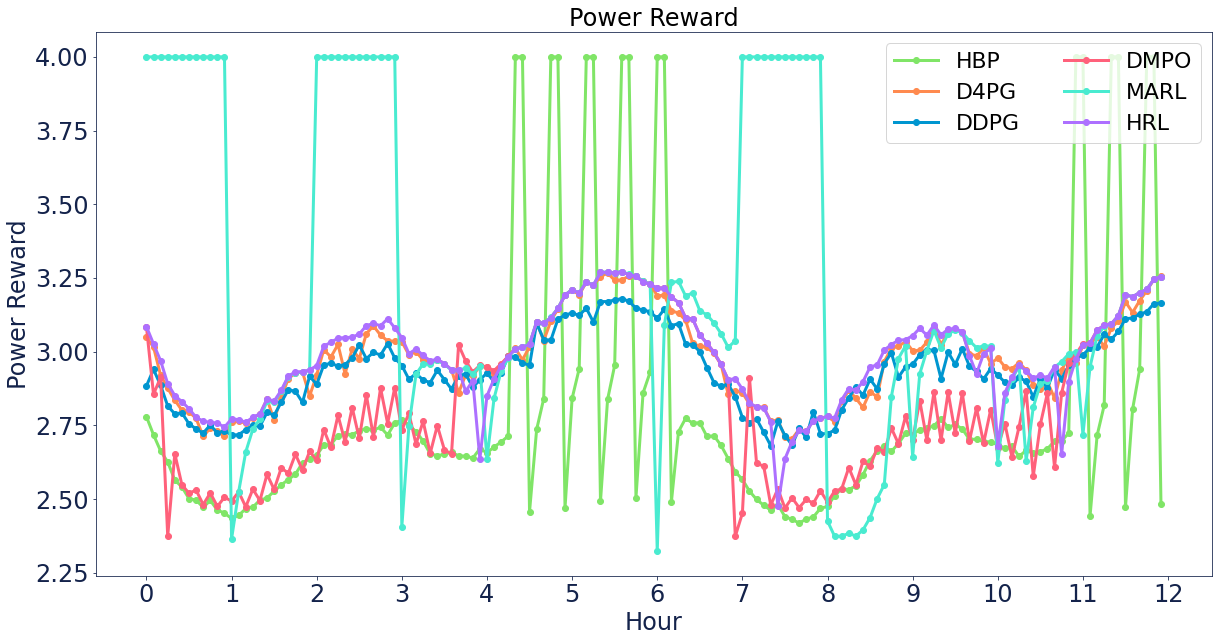} & \includegraphics[width=65mm]{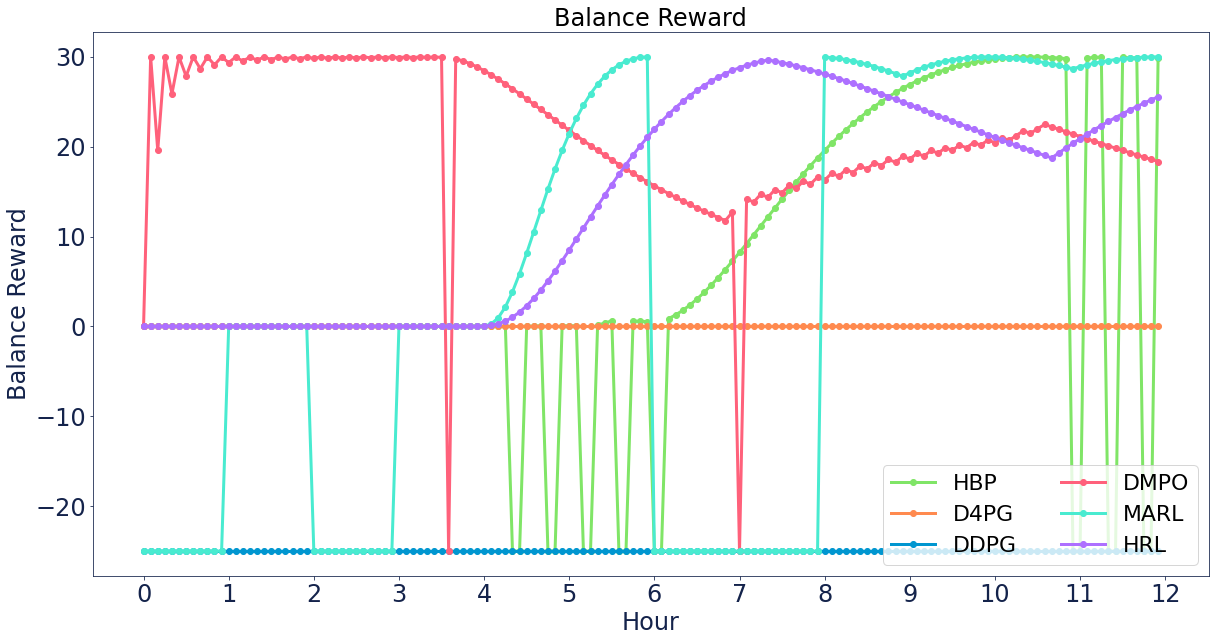}
    \end{tabular}
    \caption{Comparison of agents performance under the task reward and each component of the reward. Plots are for an arbitrary episode after training convergence to illustrate each agent's behavior. \textbf{Top left:} Task reward. \textbf{Top right:} Temperature reward. \textbf{Bottom left:} Power reward. \textbf{Bottom right:} Balance reward with the penalty requiring one chiller to be on at all times. Because entropy is 0 if only one chiller has been turned on, HRL starts with 0 balance reward. Meanwhile, DMPO tries to game this reward by turning chillers on and off frequently at the beginning of an episode. However, once HRL switches chillers after 4 hours, the agent performs better than DMPO.}
    \label{fig:rewards}
\end{figure}

\begin{figure}[htb!]
    \centering
    \begin{tabular}{cc}
        \includegraphics[width=65mm]{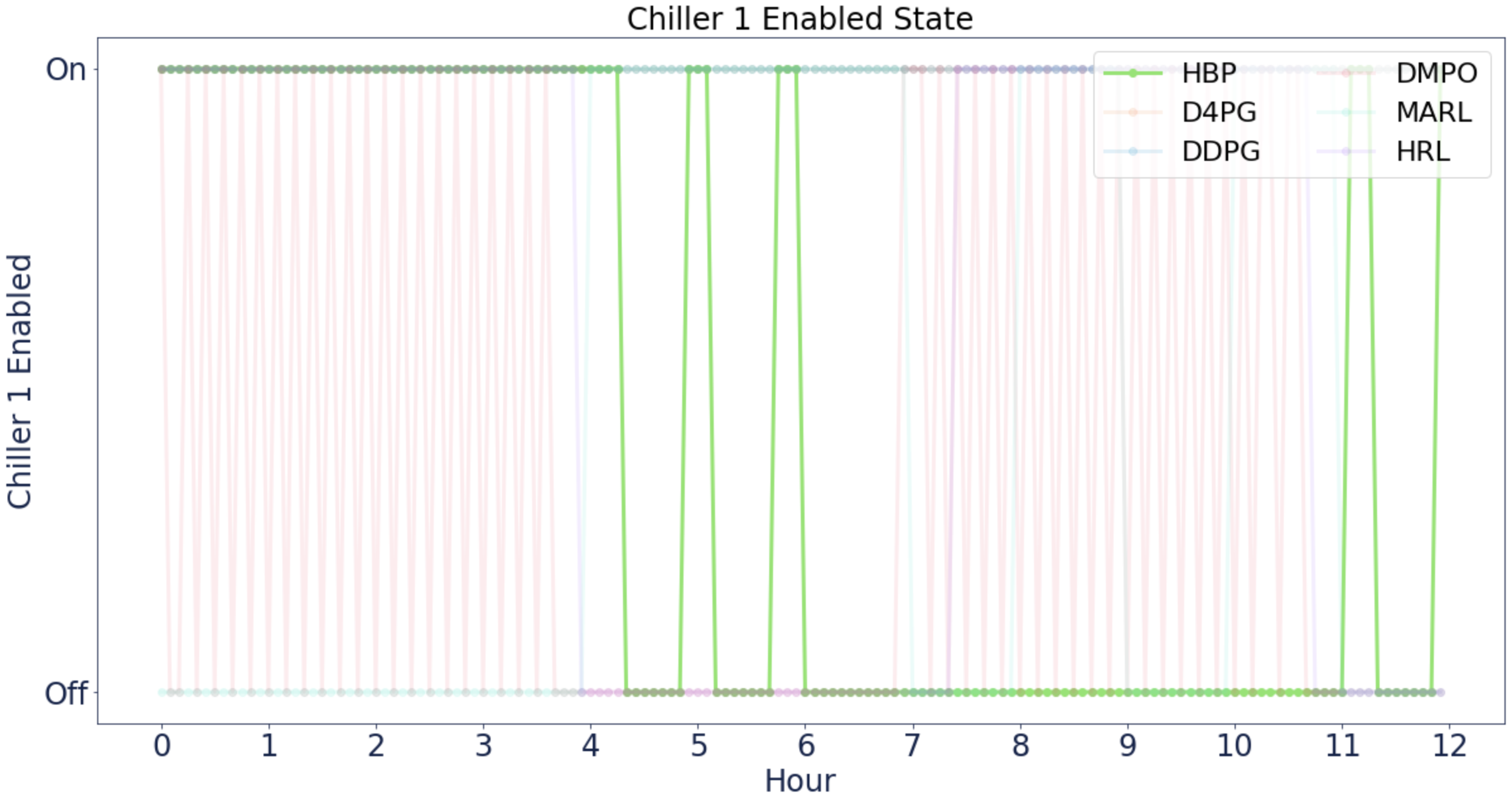} &  \includegraphics[width=65mm]{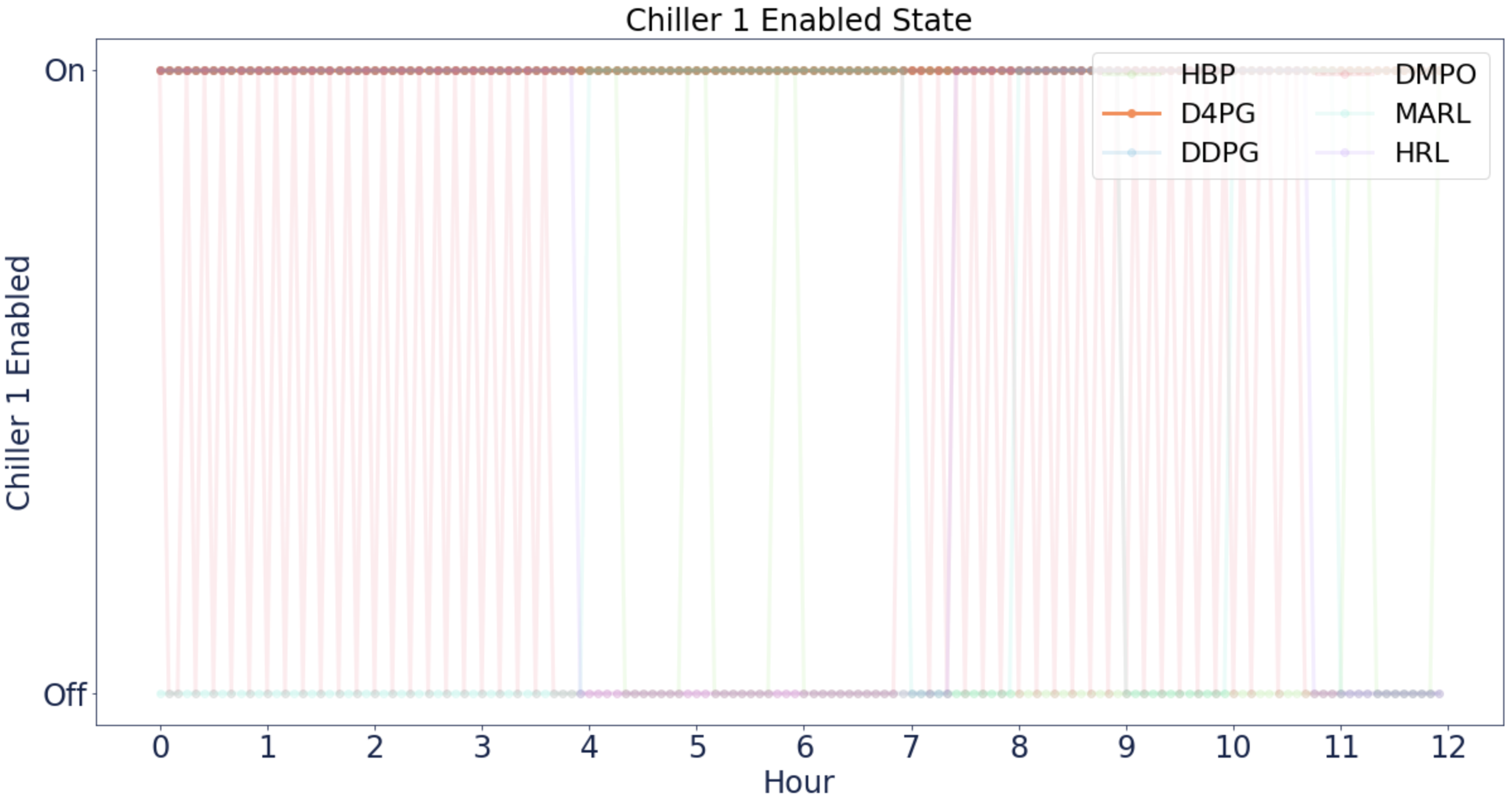}\\
        
        \includegraphics[width=65mm]{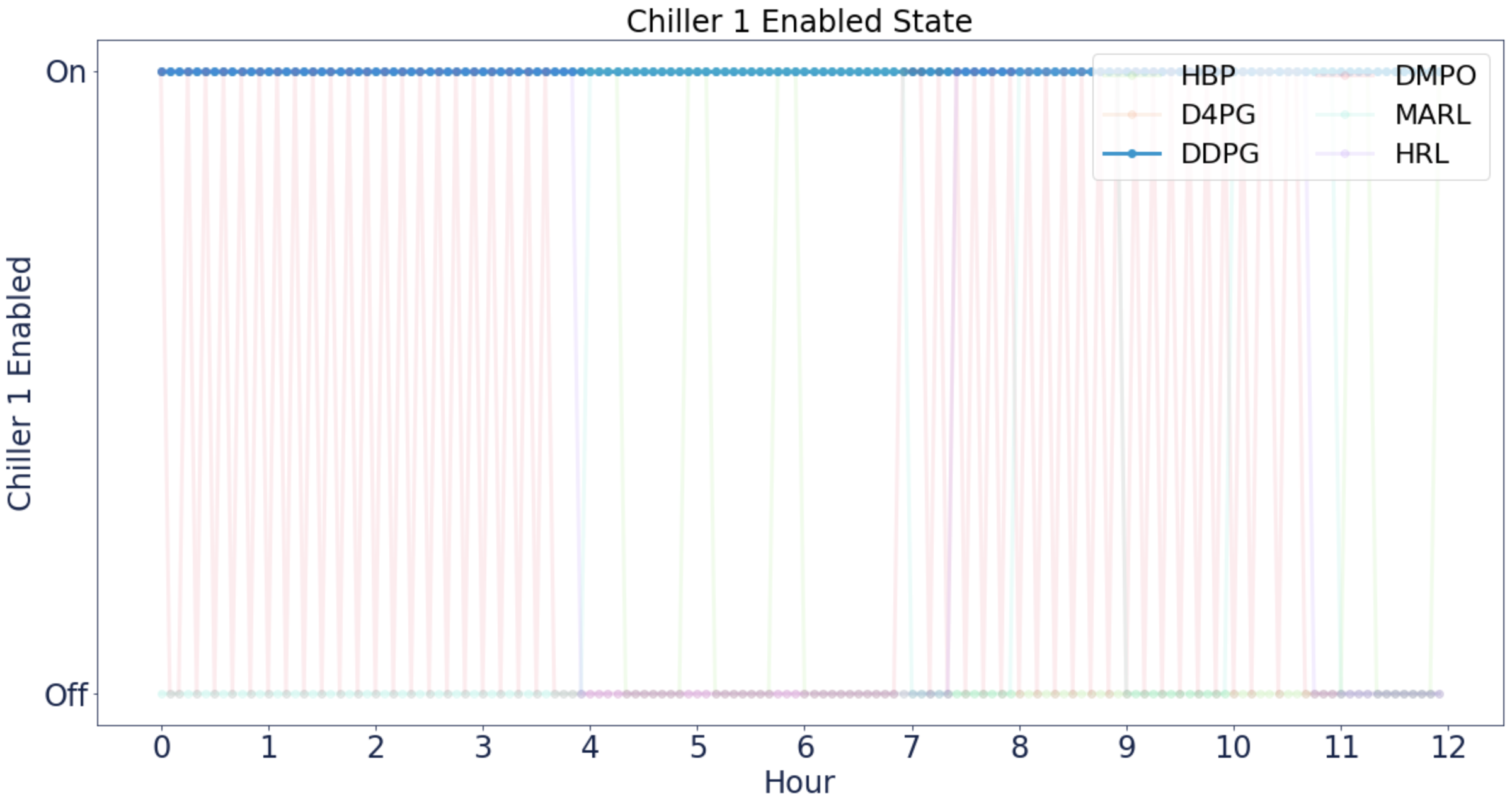} &  \includegraphics[width=65mm]{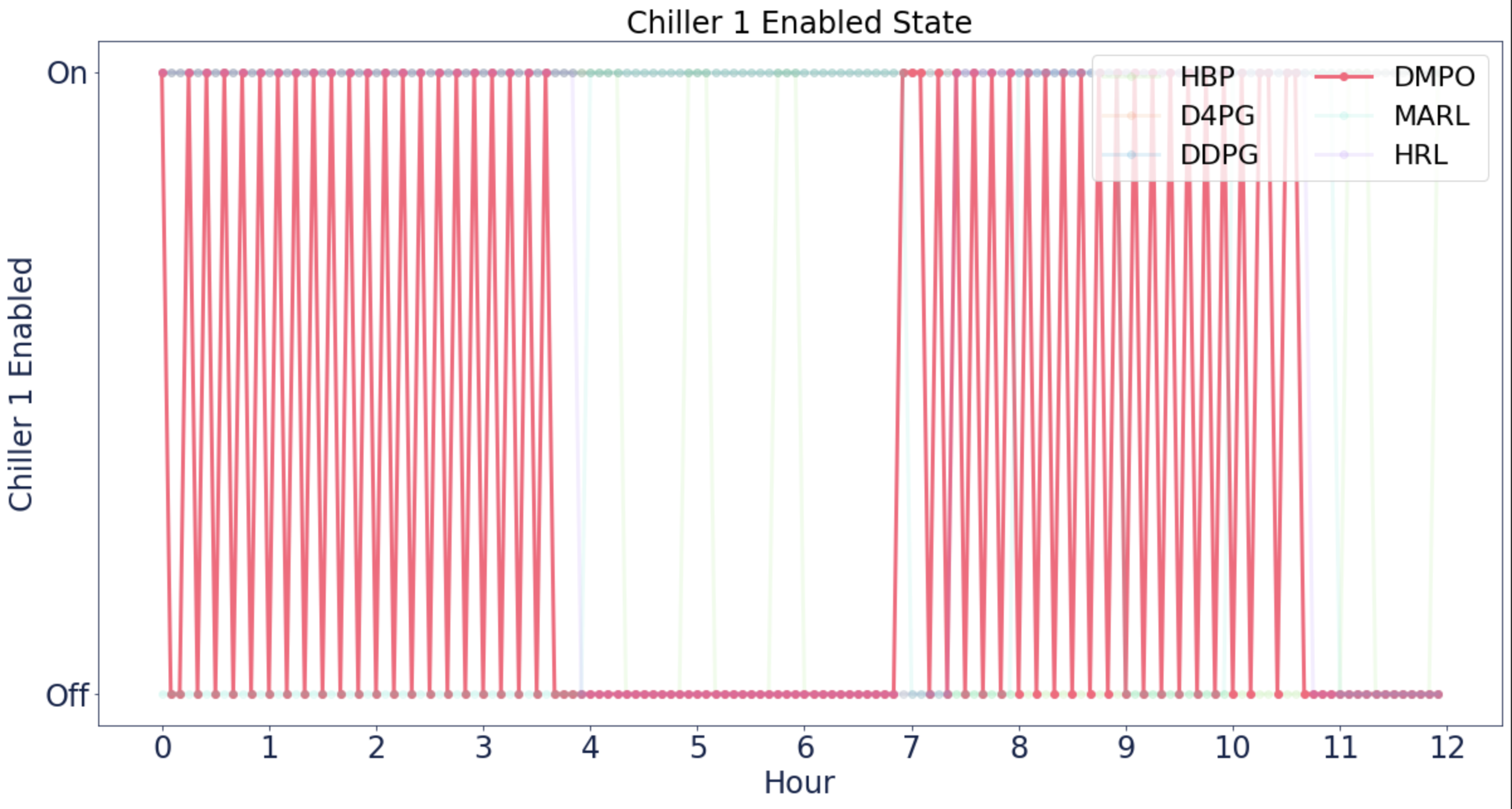}\\
        
        \includegraphics[width=65mm]{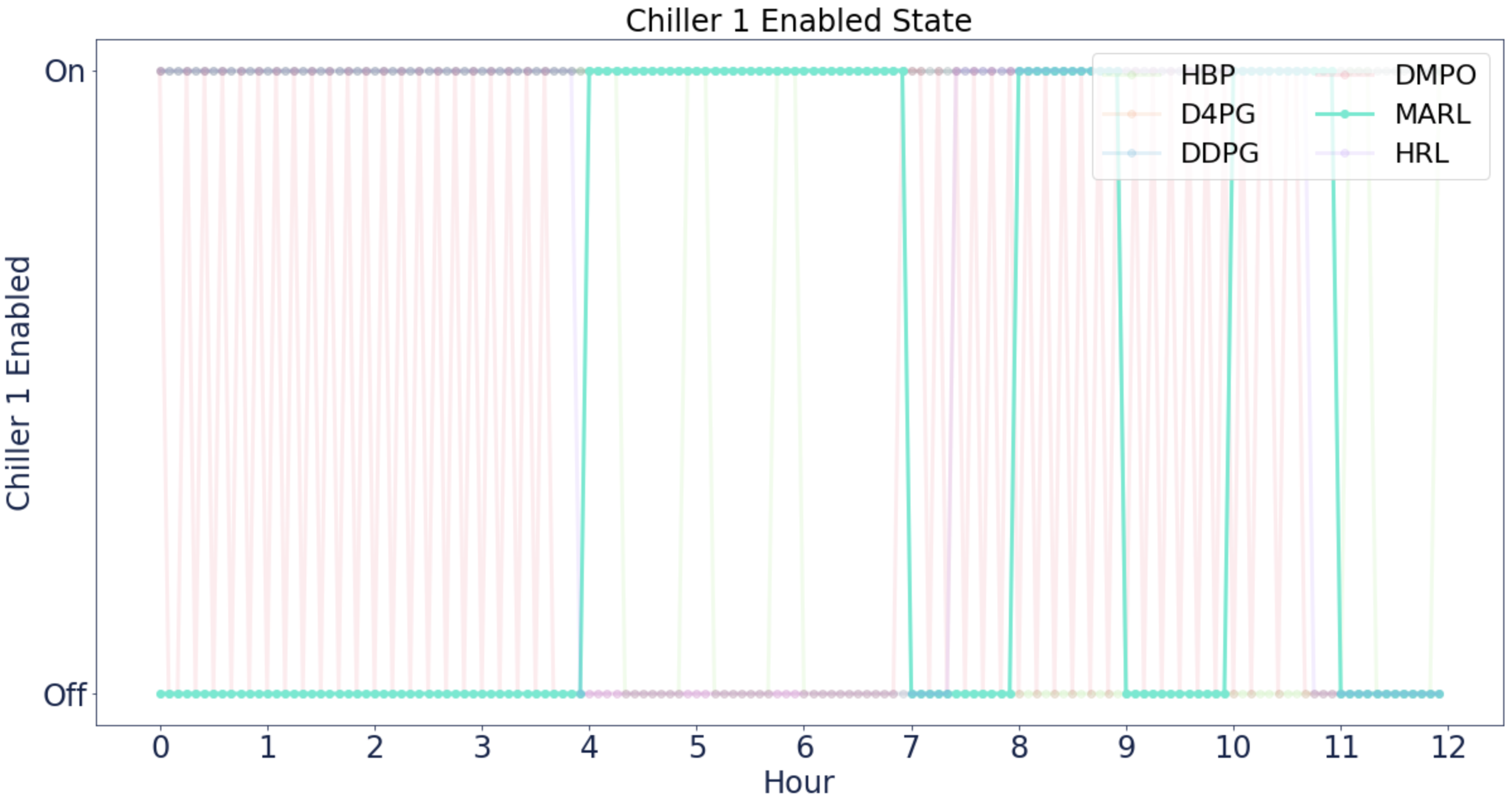} & \includegraphics[width=65mm]{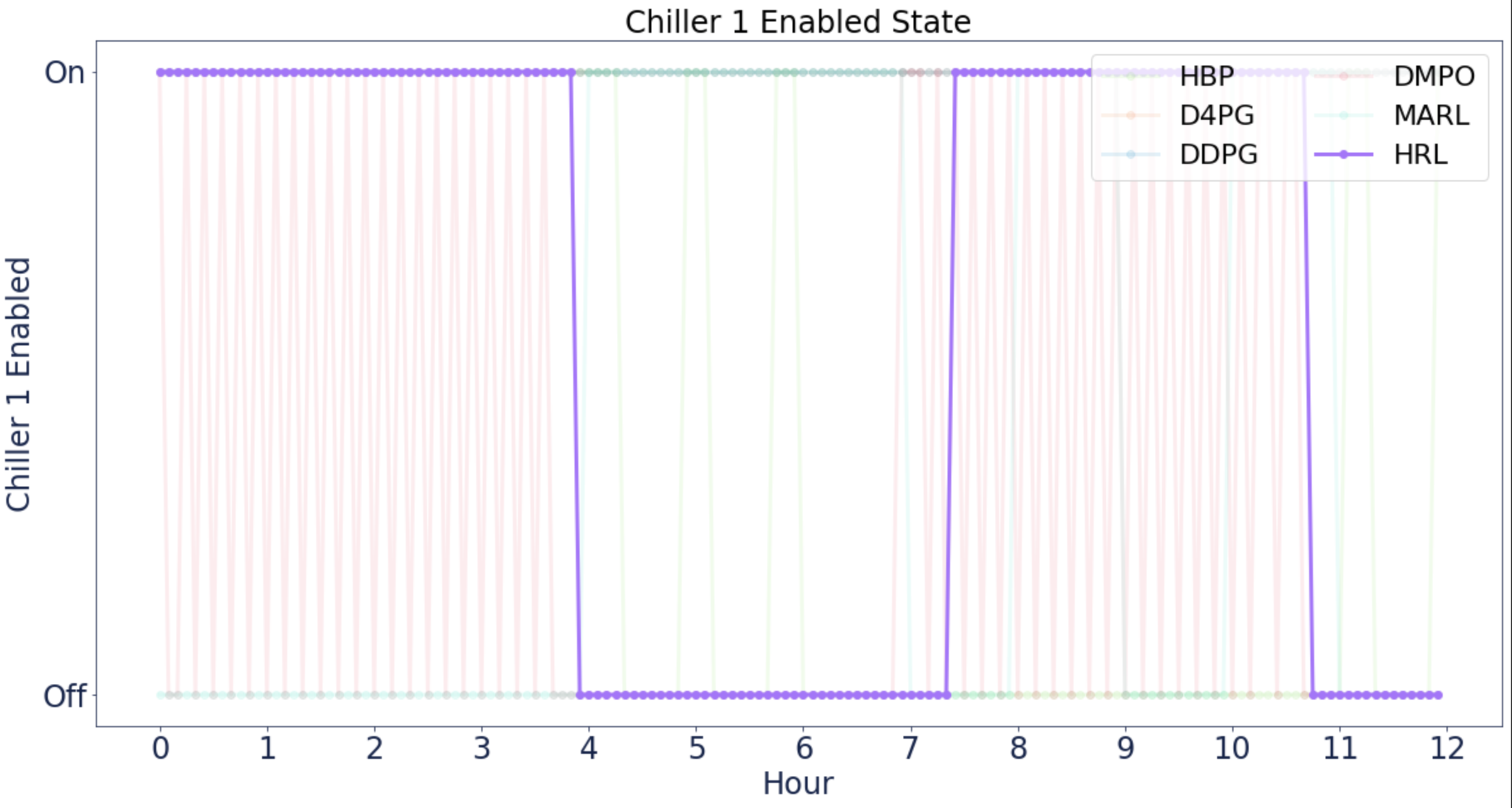}
    \end{tabular}
    \caption{Chiller 1 on and off states during an episode. HBP turns chillers on and off frequently when there is a drop in temperature, as its fixed temperature setpoint limits its ability to adapt to building load changes. D4PG and DDPG fail to learn to alternate chillers. DMPO acts unsafely, turning chillers on and off every 5 minutes. Since the MARL agent can only update chiller states once an hour, chiller states are stabler than DMPO. HRL balances chiller load \textit{and} minimizes power use the best. As turning chillers on incurs a startup energy cost, HRL reduces the number of times chillers are turned off while still utilizing both chillers equally.}
    \label{fig:chiller1_enabled}
\end{figure}
\begin{figure}[htb!]
    \centering
    \begin{tabular}{cc}
        \includegraphics[width=65mm]{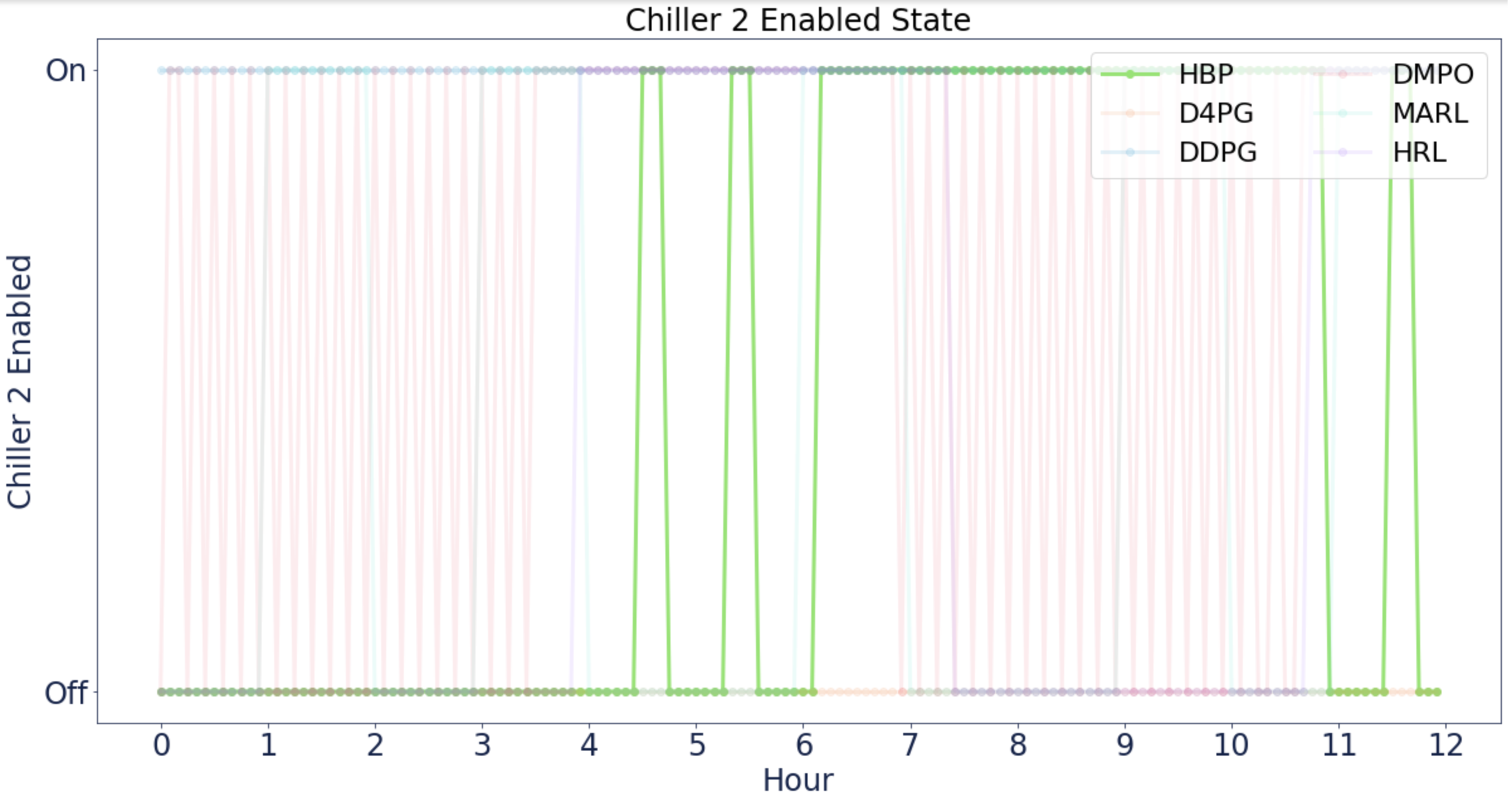} &  \includegraphics[width=65mm]{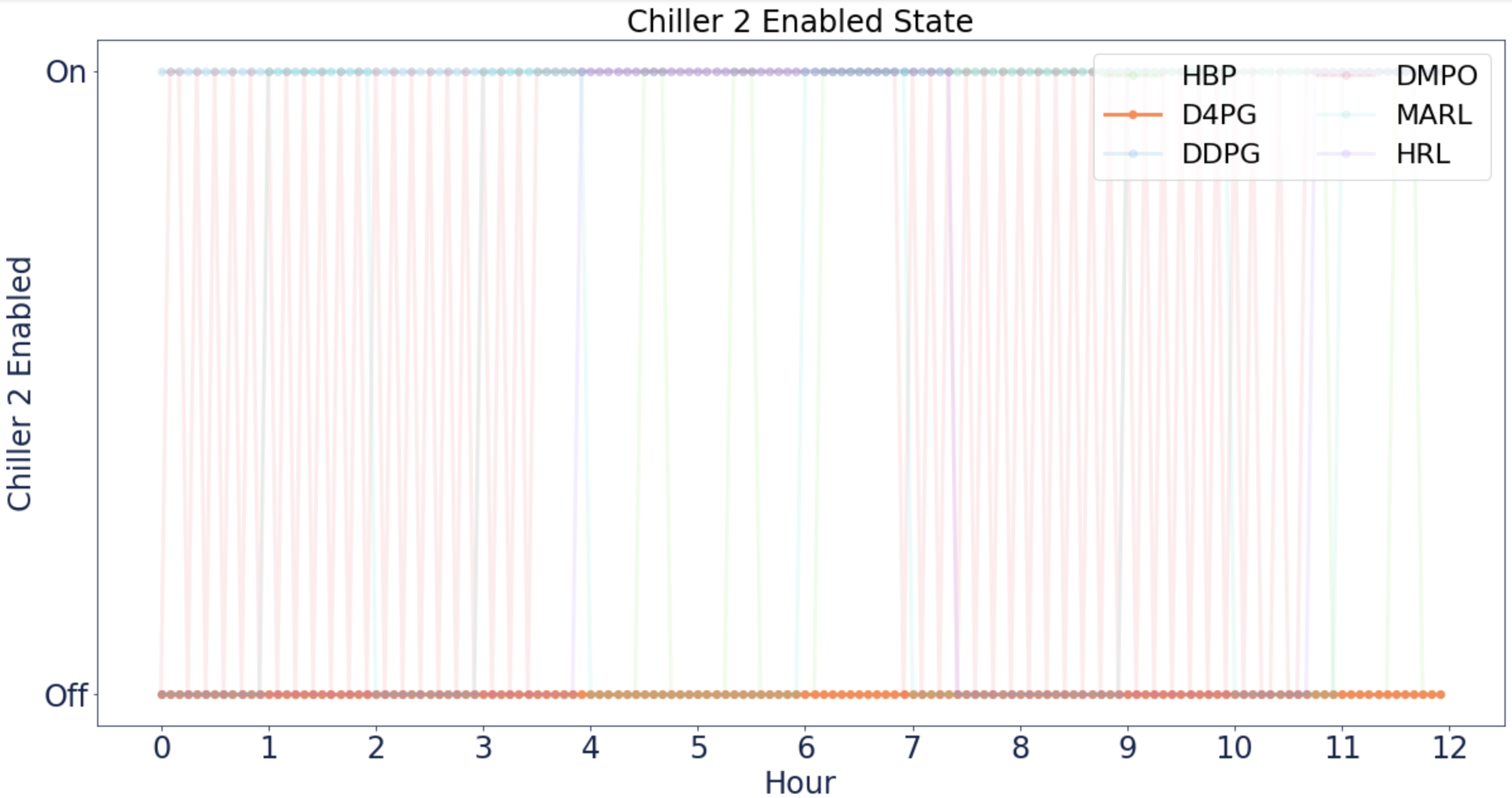}\\
        
        \includegraphics[width=65mm]{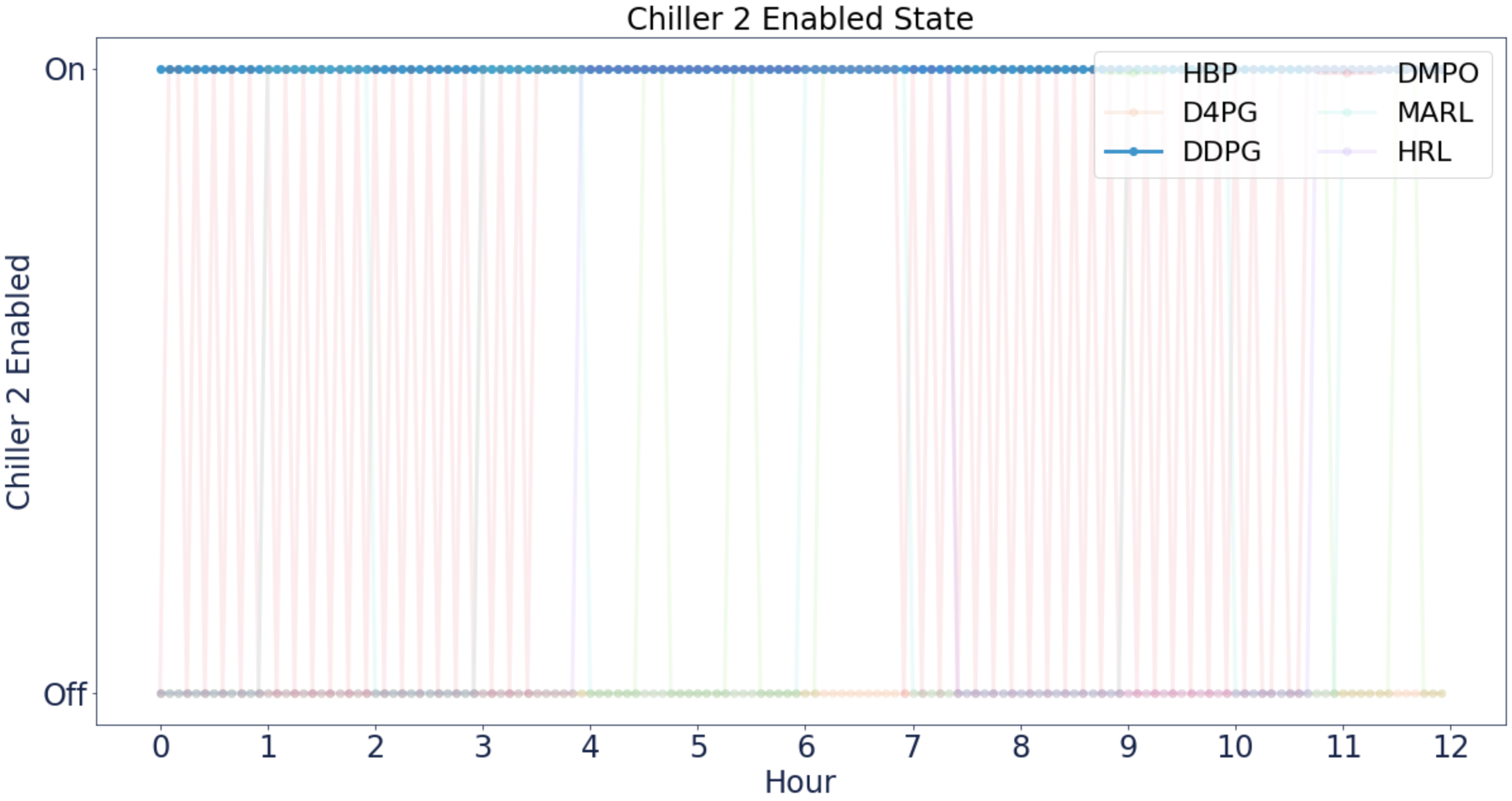} &  \includegraphics[width=65mm]{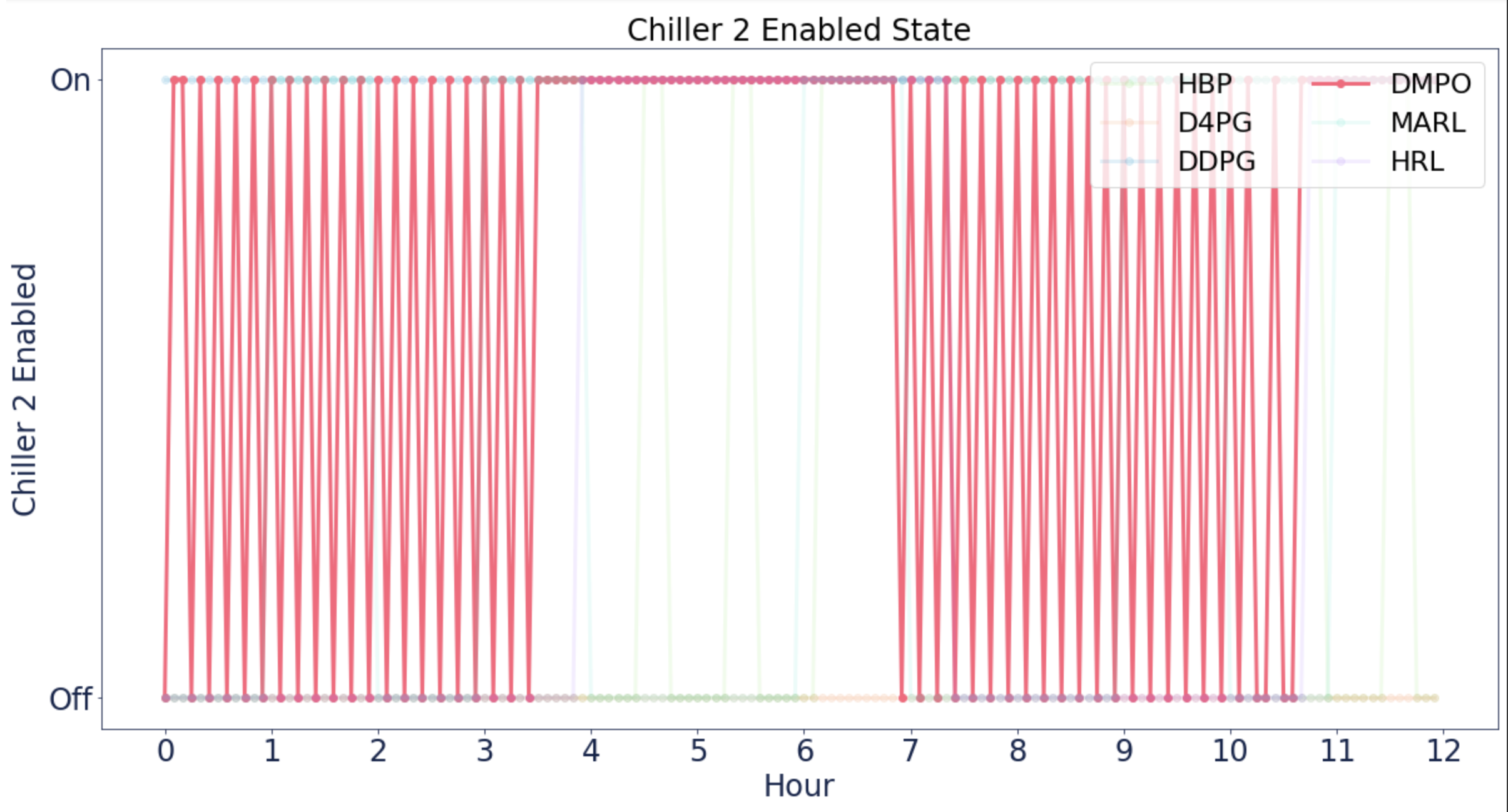}\\
        
        \includegraphics[width=65mm]{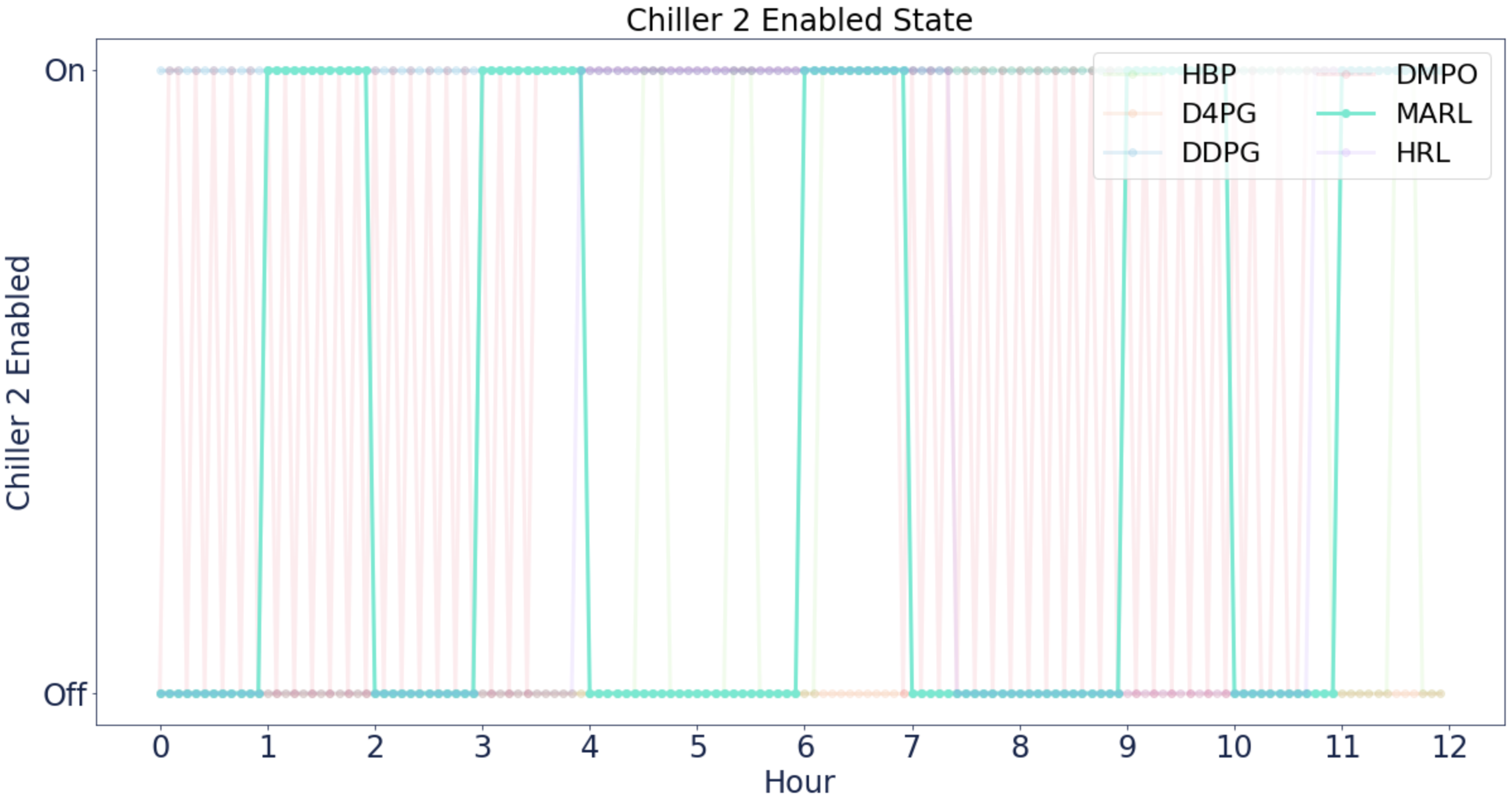} & \includegraphics[width=65mm]{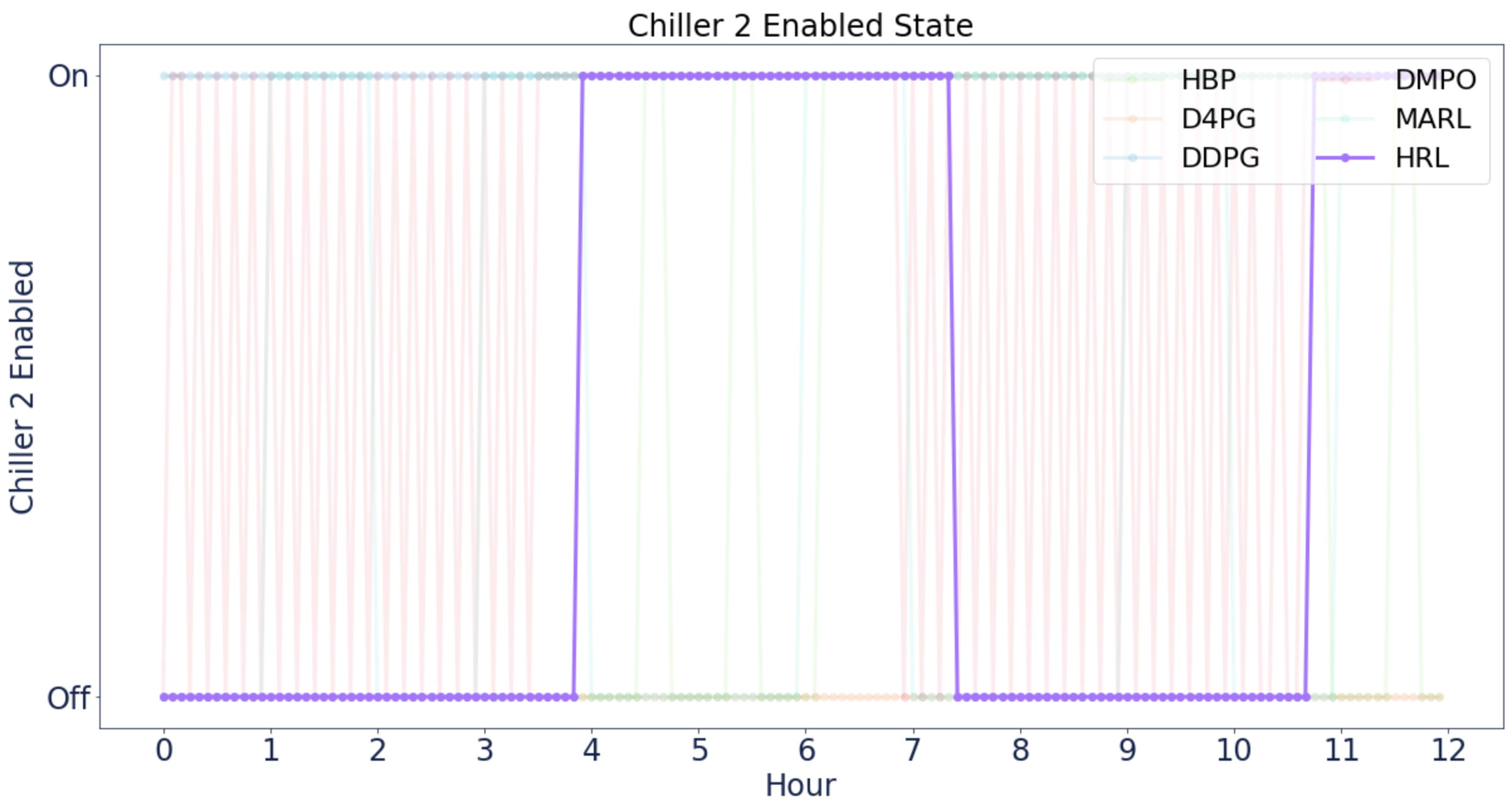}
    \end{tabular}
    \caption{Chiller 2 on and off states during an episode. Agent behaviors are similar to Figure~\ref{fig:chiller1_enabled}.}
    \label{fig:chiller2_enabled}
\end{figure}

 Training episode reward is shown in Figure~\ref{fig:episode_reward}. HRL is more sample efficient compared to DMPO, converging to a higher task reward faster. However, DMPO achieves slightly higher final reward due to the balance reward term $\alpha_{h} \cdot \textbf{h}(t)^{\lambda_{h}} - \alpha_{o} \cdot \mathbbm{1}(n_e \neq n_d)$. The entropy component is zero when only one chiller has ever been turned on. As a result, as Figure~\ref{fig:rewards} depicts, HRL achieves zero balance reward for the first 4 hours when only chiller 1 has been turned on. Meanwhile, DMPO achieves a higher balance reward during the first 4 hours since it learns that turning chillers on and off rapidly at the beginning of an episode can increase reward immediately. Figures \ref{fig:chiller1_enabled} and  \ref{fig:chiller2_enabled} depict chiller cycling behavior after training convergence for a single episode. Despite the higher task reward, the figures depict infeasible real world behavior from DMPO as rapid cycling damages chillers. As discussed in Section~\ref{sec:reward-discussion}, it is difficult to create an objective which captures all human preferences, such as infrequent cycling, and facilitates learning. Despite this, HBP and MARL avoid rapid cycling by leaving either one or both chillers on an entire episode. HRL learns the best chiller cycling behavior, using each chiller for a few hours at a time. We also observe that DDPG and D4PG learn incorrect behavior, using a single chiller or both chillers an entire episode.

\begin{figure}[htb!]
    \centering
    \includegraphics[width=90mm]{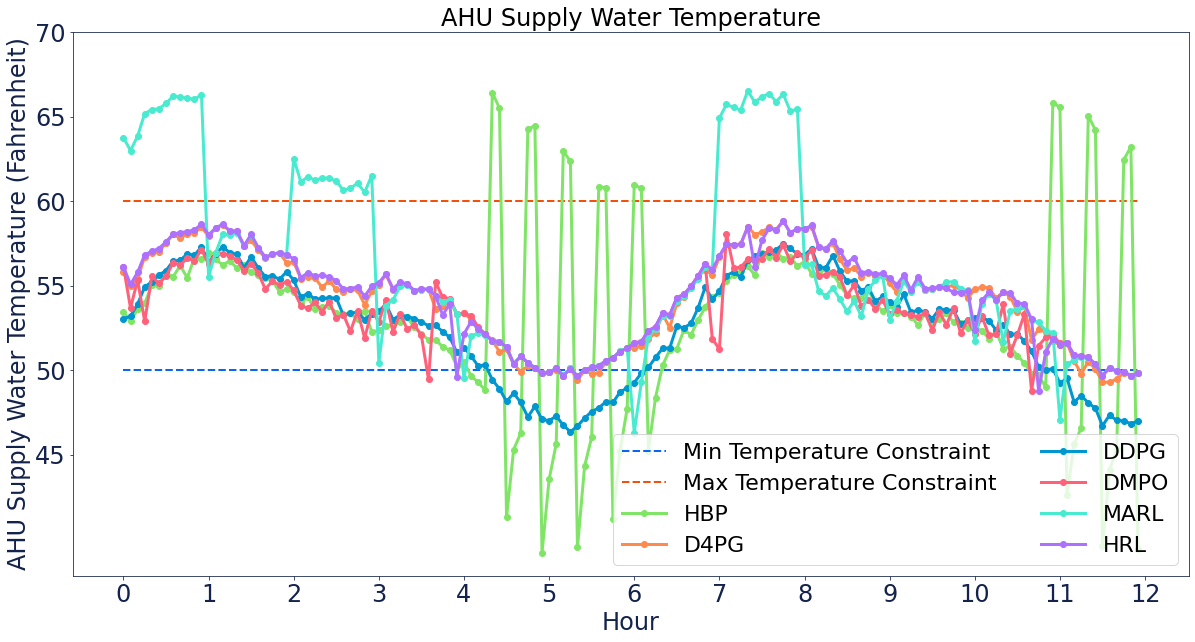}
    \caption{AHU supply water temperature achieved during an episode. DMPO and HRL learn to keep the temperature within constraints. HBP violates temperature requirements due to a fixed temperature setpoint. As a result, it switches chillers on and off to compensate for building load changes. MARL violates temperature constraints often as it does not learn that one chiller must be on at all times.}
    \label{fig:temperature_episode}
\end{figure}

\begin{figure}[htb!]
    \centering
    \includegraphics[width=90mm]{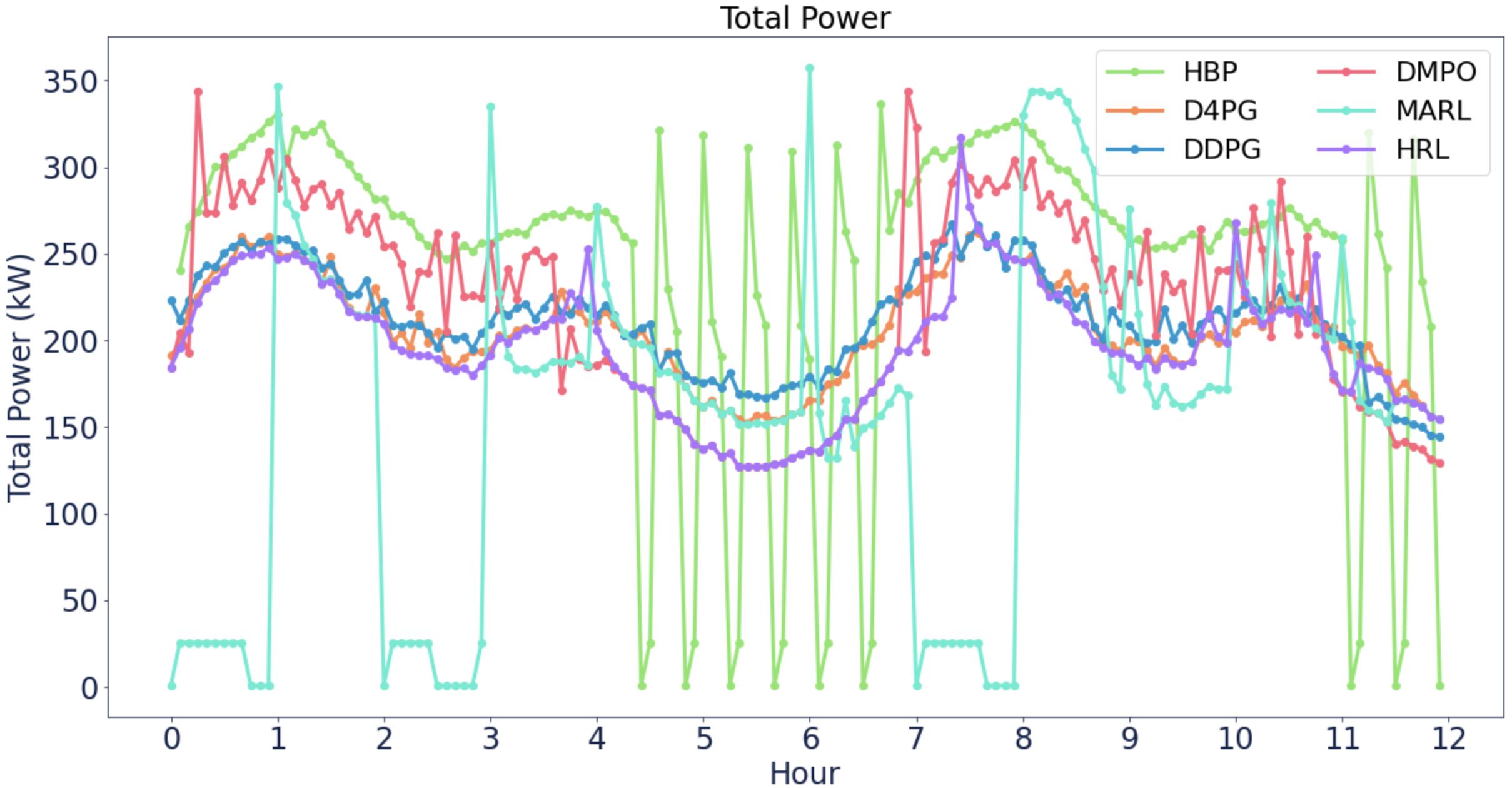}
    \caption{Turning chillers on incurs a startup cost. Since DMPO turns chillers on and off frequently, it uses more power than HRL. MARL and HBP turn chillers off to save power or meet temperature constraints, but incur high power costs when turning the chillers back on. D4PG and DDPG avoid startup costs entirely by never turning chillers off but still use more power than HRL.}
    \label{fig:power_episode}
\end{figure}

Figures \ref{fig:temperature_episode} and \ref{fig:power_episode} compare temperature and power performance respectively. Turning chillers on incurs a startup energy cost. Consequently, policies which turn chillers on and off more frequently use more energy. Across 20 random episodes with different random seeds sampled after training convergence, DMPO performs better than HRL in the balance and task reward. However, HRL uses 10\% less energy than DMPO and 17\% less energy than HBP. HBP's fixed temperature setpoint makes it difficult for the policy to meet building temperature constraints. At times, MARL uses zero chillers to save energy which causes building temperature violations. HRL and DMPO correctly use only one chiller and adjust temperature setpoints to meet temperature requirements, albeit in different ways.

\subsection{Real-World Feasibility}
\begin{figure}[htb!]
    \centering
    \includegraphics[width=70mm]{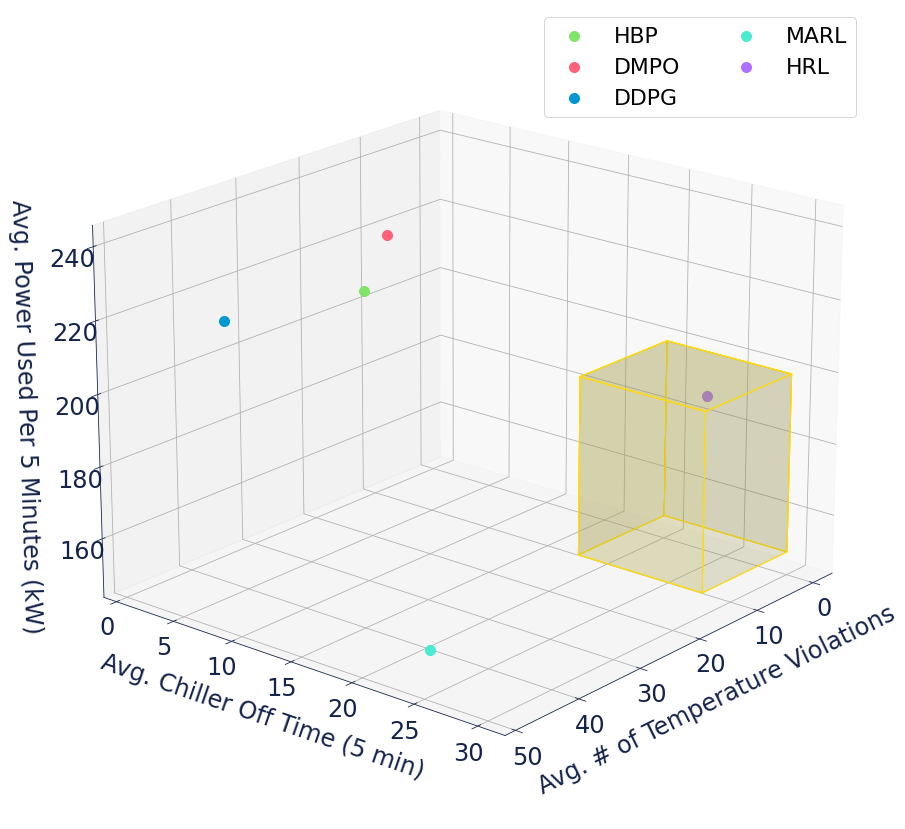}
    \caption{We simulate a real-world scenario of having limited training data by training agents on 28 days of simulation. Afterwards, we evaluate agents performance across three metrics averaged across 20 random seeds. Average number of temperature violations counts the number of times an agent violates temperature constraints. Average chiller off time measures how long each chiller is off before being switched on again during an episode. An agent which switches chillers on and off rapidly will have a low average chiller off time. The z-axis plots the average power used every 5 minutes by an agent. The \textbf{gold box} region highlights preferences for our task. We wish to minimize the number of temperature violations, have a high average chiller off time, and use less power than the HBP agent. HRL lies in this region, balancing all three objectives. \textbf{Note:} D4PG (not shown) never uses Chiller 2 resulting in an average off time off the chart.}
    \label{fig:tradeoff}
\end{figure}

Controlling HVAC systems is a safety critical task. Temperature violations can be harmful for occupants and chiller wear and tear can be costly. Additionally, simulators often are nonexistent, necessitating agents to learn from limited data. To evaluate the real-world feasibility of each algorithm, we train agents online on 28 days of simulated data then evaluate each agent's sample efficiency and number of constraint violations. As described in Section~\ref{sec:results}, HRL is more sample efficient than DMPO. To compare safety preferences, Figure~\ref{fig:tradeoff} plots each policy's average number of temperature violations across 20 episodes, average length of time each chiller is off before being turned on, and average power used every 5 minutes. A policy should minimize its temperature violations, minimize power usage, and turn chillers off for a few hours before using them again. We find that hierarchical reinforcement learning is able to balance all three objectives better than other methods.

\section{Limitations}
\label{sec:limitations}
While our work is motivated by observations of RL agent behaviour from our real life efforts, and while we showcase strong results in a simulation environment designed from real world samples, we have not yet evaluated this in a real world facility. Additionally, we showcase these results on a chiller control setup, which can be extended to a more complete HVAC problem.

\section{Conclusion and Future Work}
We present a hierarchical reinforcement learning model which  controls chillers safely and efficiently. While DMPO cycles chillers at an unsafe rate, HRL is able to use the same reward but incur less chiller wear and tear. This suggests that instead of extensive reward engineering with flat reinforcement learning, hierarchical RL offers an easier and more sample efficient alternative to achieve safe behavior. For future work, we plan to explore constrained RL and multi-objective RL as additional solutions. We also plan to extend our work to controlling an entire HVAC system.

\bibliographystyle{plainnat}
\bibliography{bibliography}


\end{document}